\documentclass[]{bytedance_seed}

\usepackage[toc,page,header]{appendix}

\usepackage{minitoc}
\usepackage{inconsolata}
\usepackage{url}
\usepackage[linesnumbered,ruled,vlined]{algorithm2e}
\usepackage{multirow}
\usepackage{booktabs}
\usepackage{caption}
\usepackage{subcaption}
\usepackage{xspace}
\usepackage{amsmath}
\usepackage{amssymb}
\usepackage{newtxmath}
\usepackage{bbm}
\usepackage{graphicx}
\usepackage{tabularx}
\usepackage{bbding}
\usepackage{pifont}
\usepackage{diagbox}
\usepackage{makecell}
\usepackage{colortbl}
\usepackage{bbm}
\usepackage{color}
\usepackage{hhline}
\usepackage{soul}
\usepackage{float}
\usepackage[most]{tcolorbox}

\newcommand{\paratitle}[1]{\vspace{1.5ex}\noindent\textbf{#1}}
\newcommand{\ie}{\textit{i.e.,}\xspace}

\newcommand{\eg}{\textit{e.g.,}\xspace}

\newcommand{\ignore}[1]{}

\definecolor{takeaway}{RGB}{165, 209, 216}
\definecolor{takeawayTitle}{RGB}{57, 89, 163}

\title{ Adaptive Ability Decomposing for Unlocking Large Reasoning Model Effective Reinforcement Learning }

\author[1,2,*]{Zhipeng Chen}
\author[2]{Xiaobo Qin}
\author[1, \dagger]{Wayne Xin Zhao}
\author[2, \dagger]{Youbin Wu}
\author[1]{Ji-Rong Wen}

\affiliation[1]{Renmin University of China}
\affiliation[2]{ByteDance Seed}

\contribution[*]{Work done at ByteDance Seed}
\contribution[\dagger]{Corresponding authors}

\abstract{
Reinforcement learning with verifiable rewards (RLVR) has shown great potential to enhance the reasoning ability of large language models (LLMs).
However, due to the limited amount of information provided during the RLVR process, the model can only engage in largely blind exploration, which often results in failure on challenging problems.
To provide additional information for the RLVR process without relying on a teacher model, we propose \textbf{A$^2$D}, an \textbf{A}daptive \textbf{A}bility \textbf{D}ecomposing method for enhancing the effectiveness of RLVR.
Specifically, we first train a decomposer via RLVR without distillation, enabling it to decompose complex questions into a set of simpler sub-questions.
Next, we use this decomposer to annotate sub-questions for each question in the training dataset, and then train the reasoner under RLVR with sub-question guidance.
To better understand A$^2$D, we first compare its performance with competitive baselines, showing its effectiveness.
Next, we observe that our method functions as a plug-and-play module that can be applied to different RLVR algorithms.
Furthermore, we conduct an analysis of the decomposer, revealing how the RLVR process affects its performance and behavior, and which type of guidance is better suited for enhancing the reasoner's exploration and exploitation abilities.
}

\date{\today}
\correspondence{Zhipeng Chen at \email{zhipeng\_chen@ruc.edu.cn}, Wayne Xin Zhao at \email{batmanfly@gmail.com}, Youbin Wu at \email{youbinwu@bytedance.com}}

\begin{document}
\maketitle

\vspace{-25pt}
\begin{figure}[h]
    \centering
    \includegraphics[width=1.0\linewidth]{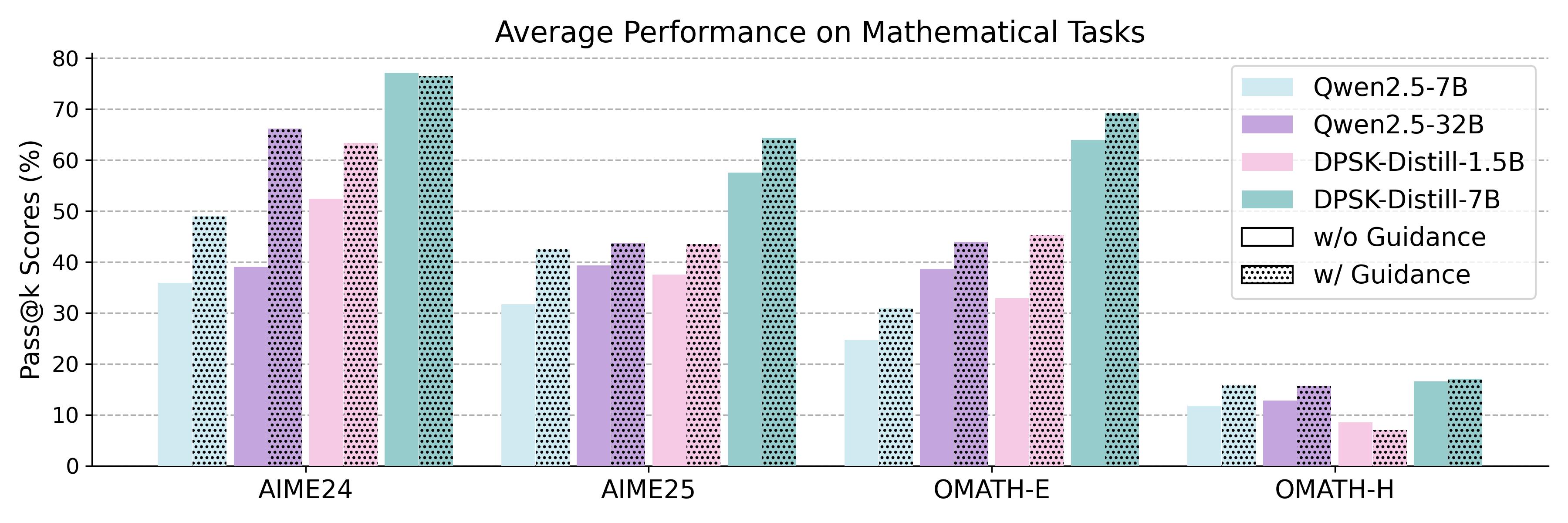}
    \caption{Average performance of various models on mathematical tasks. By incorporating sub-question guidance into the prompts, LLMs can perform exploration and reasoning more effectively, thereby achieving higher Pass@k scores.}
    \label{fig:overview}
\end{figure}

\newpage

\section{Introduction}
\label{sec:intro}

Recently, reinforcement learning with verifiable rewards (RLVR) has become a widely used approach to enhance the reasoning ability of large language models (LLMs), \eg OpenAI o3~\cite{openai_o3}, DeepSeek-R1~\cite{deepseek-r1}, and Seed1.5-VL~\cite{seed1.5-vl}.
Through the RLVR training procedure, the model can possess the ability to perform complex reasoning actions~\cite{still-3,open_reasoner_zero}, \eg self-reflection, self-verification, and self-correction, which can also be called the large reasoning model (LRM).
During the RLVR training process, the LRMs first explore and reason over a given question, and then a verifier provides rewards for the generated solutions. 
The LRMs learn from these experiences and gradually acquire the ability to perform complex reasoning behaviors~\cite{kimi-k1.5,seed1.5-thinking}.

Although RLVR has achieved substantial success in effectively improving the reasoning capabilities of LRMs, a series of studies~\cite{Yue-arxiv-2025-Does,Yao-arxiv-2025-TheDebate,Nguyen-arxiv-2025-TheReasoning,Wu-arxiv-2025-TheInvisible} have pointed out that RLVR struggles to raise the upper bound of model performance, thereby hindering continued capability growth in later training stages.
To alleviate this issue, prior work~\cite{entropyadv,passk_training,pkpo,Gai-arxiv-2025-differential} has focused on designing reward functions and advantage functions to encourage LRMs to explore and solve more challenging problems, thereby enhancing the model’s exploration capability.
Despite encouraging the model to explore, supervision provided only at the outcome level, which is the popular RLVR training settings~\cite{grpo,deepseek-r1}, offers limited guidance for effective exploration~\cite{on_policy_distillation}, leaving the model in a state of largely blind exploration~\cite{Song-arxiv-2025-outcome,Wang-arxiv-2025-masked,RLVMR}.
In this case, the model’s exploration often leads to failure in solving the given problems, which will be punished by the RLVR process, causing the model to adopt conservative reasoning strategies and reduce further exploration.
In the RLVR procedure, which is a \textit{low-information setting}~\cite{liu-arxiv-2025-stabilizing,Castanyer-arxiv-2025-Stable}, the loss of the model’s exploration capability implies that further training is unlikely to yield performance improvements~\cite{Yue-arxiv-2025-Does,Dong-arxiv-2025-RLPLUS,Wang-arxiv-2025-Reinforcement}, limiting the scaling process.

Existing studies~\cite{on_policy_distillation} have discussed the amount of information provided to the model by different training methods, noting that RLVR can only supply less information than the distillation process.
Following this theory, to increase the amount of information available during RLVR training, one can introduce distilled data or manually annotated data.
We consider whether decomposing the original problem into subproblems and providing them as additional information to the model can further raise its performance upper bound, since we observe that some prior work suggests that planning then reasoning can achieve better performance~\cite{plan_tuning,plan_and_act}. 
To verify this hypothesis, we conduct the empirical study and present the results in Fig.~\ref{fig:overview}.
The model achieves higher Pass@k scores with subquestion-based prompts, indicating that the model already possesses the requisite knowledge to solve the task and can effectively integrate this knowledge to reach a solution when guided appropriately.
This phenomenon suggests that the model lacks compositional generalization, \ie the ability to combine simple knowledge components to solve complex problems~\cite{Liu-nips-2020-Compositional,An-acl-2023-How}, demonstrating the effectiveness of incorporating additional information into RLVR and sub-question guidance can serve as one of the effective forms of additional information.

To provide additional information to the RLVR procedure, prior work (\eg LUFFY~\cite{luffy} and Scaf-GRPO~\cite{scaf-grpo} has guided the model to perform more effective exploration during RLVR training by introducing data generated by a powerful model (\eg DeepSeek-R1) as prompts, including knowledge point, planning, and solution.
These methods rely on stronger models to annotate data, which on one hand incurs higher costs, and on the other hand, they remain within the paradigm of imitation learning, making it difficult for the student model to surpass the upper bound of the teacher model~\cite{TAID,Trinh-arxiv-2025-GUIDE}.
According to the above discussion, we consider whether it is possible to start from a single model and, by employing different tasks and RLVR training methods, derive a hint model and a reasoning model. 
Based on a given problem, the hint model can supply additional information to assist the reasoning model in performing more effective exploration.
In this setting, the decomposer and reasoner trained from the same model can mutually reinforce each other, thereby enabling model self-evolution and surpassing its reasoning performance limits.

Building on this idea, in this paper, we propose an \textbf{A}daptive \textbf{A}bility \textbf{D}ecomposing method for enhancing the effectiveness of RLVR, named as \textbf{A$^2$D}.
Concretely, we first utilize RLVR to train a decomposer that can serve as the hint model and decompose the original questions into simple sub-questions, providing additional information for training the reasoning model, since decomposing a problem can help the model solve the complex original tasks~\cite{Zhou-arxiv-2025-AnApproach,Qasim-journal-2025-Recursive}.
Next, differing from earlier studies that simply insert prompt-generated answers into the training dataset~\cite{luffy,scaf-grpo}, we utilized \textit{in-context distillation loss} (IDL) to guide the model to learn from the experiences obtained through sub-question-guided exploration.
This makes our algorithm a plug-and-play component, capable of adapting to various RLVR algorithms.

In summary, the major contributions of our work can be summarized as follows:
\begin{itemize}
    \item To provide more information without relying on external models, we propose A$^2$D, training a decomposer and a reasoner starting from the same model. The decomposer is used to decompose questions, and the decomposed sub-questions serve as additional information to assist the reasoner during RLVR.
    \item We evaluate and compare our A$^2$D with competitive baselines on eight mathematical tasks to validate the effectiveness. With the assistance of sub-question guidance from the decomposer, the reasoner is able to explore more effectively during various RLVR algorithms and achieve better performance.
    \item We analyze the decomposer and its training process, revealing how different methods affect its performance and behavior. We also find that specific and fine-grained hints help improve the reasoner’s Pass@1 score, while abstract and coarse-grained hints can enhance the reasoner’s Pass@k score.
\end{itemize}
\section{Approach}
\label{sec:approach}

\begin{figure}[t]
    \centering
    \includegraphics[width=1\linewidth]{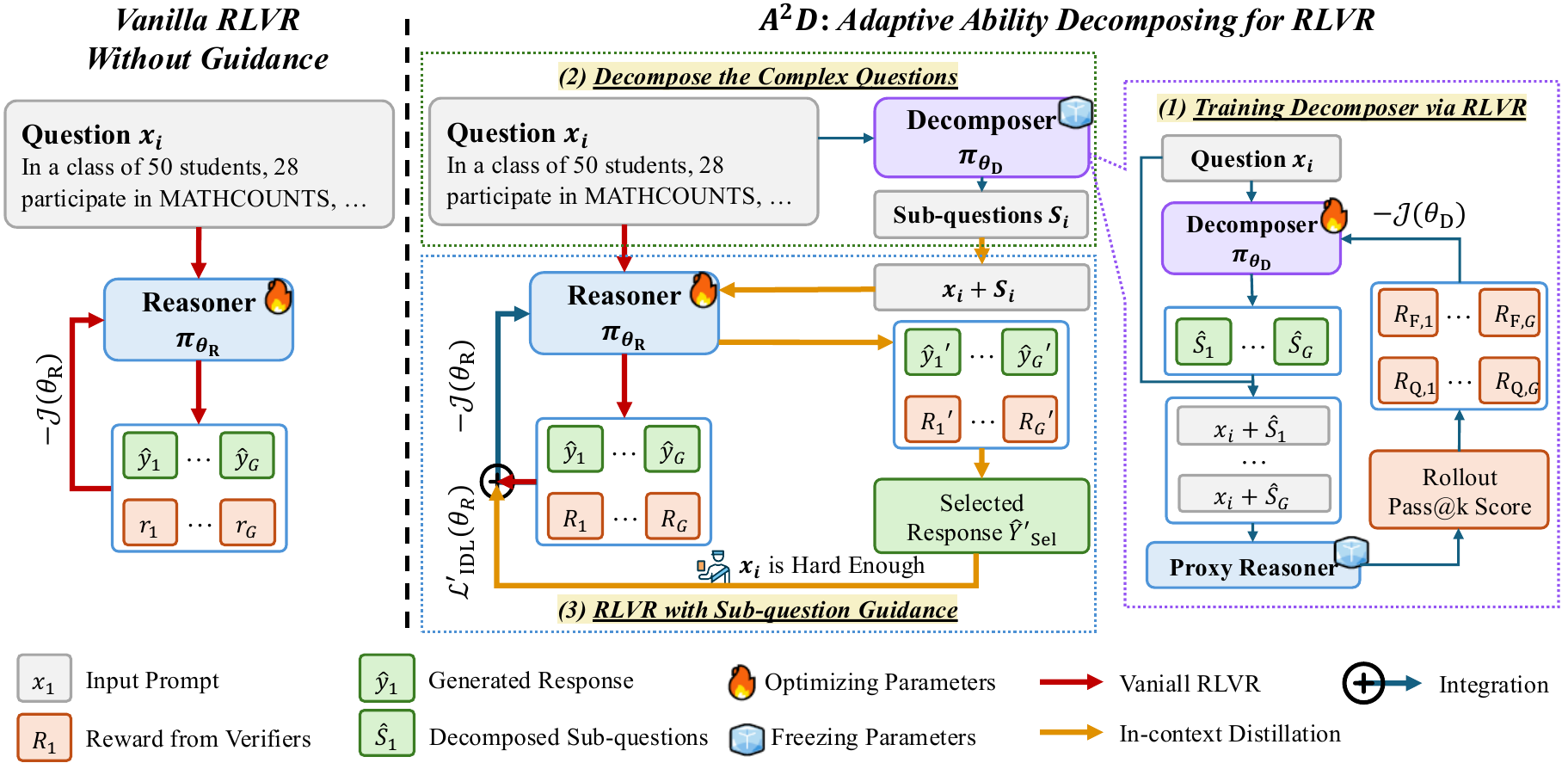}
    \caption{The comparison between vanilla RLVR and our approach A$^2$D. In A$^2$D, we first train a decomposer $\pi_{\theta_\text{D}}$ through RLVR, guided by the format reward $R_\text{F}$ and quality reward $R_\text{Q}$. Next, we use this decomposer to annotate sub-questions for each question in the training dataset, and design the in-context distillation loss (IDL), \ie Eq.~\ref{eq:idl},  to integrate the sub-question guidance into the RLVR process for reasoner $\pi_{\theta_\text{R}}$. To better demonstrate the pipeline of our A$^2$D, we present the pseudo code in Algorithm~\ref{code:a2d}.}
    \label{fig:framework}
\end{figure}

In this section, we first review the popular RLVR process (Sec.~\ref{sec:preliminary}). Next, we introduce our approach A$^2$D, including decomposer training (Sec.~\ref{sec:train_decompoer}) and RLVR with guidance (Sec.~\ref{sec:train_policy_model}).
Concretely, in A$^2$D, we first utilize reinforcement learning to train a decomposer $\pi_{\theta_\text{D}}$, which can decompose the complex question into several sub-questions that LLMs can easily solve. 
Next, we leverage it to generate the sub-questions of each question from the training dataset as the corresponding guidance.
After that, the guidance can be employed to improve the effectiveness of the RLVR process to train a reasoner $\pi_{\theta_\text{R}}$, which can solve the downstream tasks better.
We present the overview of A$^2$D in Fig.~\ref{fig:framework}.

\subsection{Preliminary: Reinforcement Learning with Verifiable Reward}
\label{sec:preliminary}
Reinforcement learning with verifiable reward (RLVR) is a popular algorithm to further enhance the reasoning abilities of LLMs~\cite{openai-o1,deepseek-r1,kimi-k1.5}, \eg GRPO~\cite{grpo}, RLOO~\cite{rloo}, and REINFORCE++~\cite{reinforce++}.
In the RLVR training dataset $\mathcal{D}=\{(x_i,y_i)\}_{i=1}^{n}$, the $i$-th training instance contains a question description $x_i$ and a ground truth answer $y_i$.
Give the question $x$, a policy with the parameter $\theta$ (denoted as $\pi_\theta$) is required to generate a solution with a final answer $\hat{y}$.
Once the predicted answer $\hat{y}$ is obtained, a verifier will be utilized to verify its correctness and provide the reward $r$ based on the ground truth answer $y$.
When the predicted answer is correct, the verifier provides a positive reward $R_\text{pos}$, while it provides a negative reward $R_\text{neg}$ for other situations.
Based on question $x$, generated solution and answer $\hat{y}$, and reward $r$ from the verifier, following the process of GRPO, the objective function of the RLVR can be formulated as follows,
\begin{equation}
\small
\label{eq:grpo_loss_function}
\mathcal{J}(\theta)=\mathbb{E}_{(x,y)\sim\mathcal{D},\{\hat{y}\}_{i=1}^G\sim\pi_\theta(\cdot|x)}\left[\frac{1}{G}\sum_{i=1}^{G}\frac{1}{|\hat{y}_i|}\sum_{t=1}^{|\hat{y}_i|}\min\left(r_{i,t}\hat{A}_{i,t},\text{clip}\left(r_{i,t},1-\varepsilon, 1+\varepsilon\right)\hat{A}_{i,t}-\beta D_\text{KL}\right)\right],
\end{equation}
where $r_{t}$ and $\hat{A}_t$ refer to the importance sampling coefficient and estimated advantage values, and $G$ denotes the number of generated responses based on question $x$.
In GRPO, the average value and variance of the rewards in the same group are leveraged to estimate the advantage value as the following equation,
\begin{equation}
\small
\label{eq:grpo_advantage}
\hat{A}_{i,1}=\dots,\hat{A}_{i,|y_i|}=\hat{A}_i = \frac{R_i - \text{mean}(R_1,\dots,R_G)}{\text{std}(R_1,\dots,R_G)},
\end{equation}
where $\text{mean}(R_1,\dots,R_G)$ and $\text{std}(R_1,\dots,R_G)$ denote the average value and variance of the rewards in the corresponding group, respectively.
Moreover, each token in response $\hat{y}_i$ will receive the same advantage $\hat{A}_i$.

\subsection{Training Decomposer via RLVR}
\label{sec:train_decompoer}
Given the large consumption of data annotating for the SFT process of the decomposer, we consider utilizing RLVR to guide the LLMs to learn to decompose the complex question through self-exploration.
In the following, we first introduce the construction of the training dataset of the decomposer $\pi_{\theta_\text{D}}$, and then present the details of the RLVR process.

\paratitle{Training Dataset Construction.}
We collect the question and its corresponding answer to construct an instance of the training dataset.
Formally, we construct a training dataset $\mathcal{D}=\{\langle x_i,y_i \rangle\}_{i=1}^n$, where $x_i$, $y_i$, and $n$ denote the question, the corresponding ground truth answer, and the number of training data.
As discussed in previous work~\cite{tot,ADaPT}, decomposing the original question is essentially a planning process, during which the model possesses a greater capacity to explore potential solution strategies.
Under this assumption, we think that the trained decomposer can provide useful guidance for the subsequent training of the reasoner, even though no additional information is introduced throughout the entire training process.

\paratitle{RLVR Training Procedure.}
To effectively train the Decomposer, we design a multi-dimensional reward function, including both a quality reward $R_\text{Q}$ and a format reward $R_\text{F}$.
Concretely, to evaluate the quality of the sub-questions generated from decomposer, we conduct a proxy reasoner to generate the solution for $n_\text{proxy}$ times based on the original question and the decomposed sub-questions.
The quelity reward is set to $1$ if and only if at least one of the $n_\text{proxy}$ attempts is correct (\ie $R_\text{Q}=1$).
Otherwise, the quality reward is set to 0 (\ie $R_\text{Q}=0$).
Moreover, for the format reward, we examine whether the responses from the decomposer satisfy the required format, \ie whether it uses ``<subquestion>'' to indicate the beginning of the decomposed sub-questions and whether the corresponding content is non-empty (The details can be found in Appendix~\ref{app:details_format_reward}).
The format reward will be set to $1$ when and only when the response passes the format examination (\ie $R_\text{F}=1$), while it will be set to $0$ in other situations (\ie $R_\text{F}=0$).
After obtaining the quality reward and format reward, we compute the final reward for a response from the decomposer by multiplying $R_\text{Q}$ and $R_\text{F}$, \ie $R=R_\text{Q}\times R_\text{F}$.
With this reward design, we encourage the model to produce responses in the correct format and to assist the proxy reasoner in generating correct answers, thereby reducing the likelihood of reward hacking.
In practice, we utilize GRPO~\cite{grpo} to train the decomposer based on the above reward function.

\subsection{Improving RLVR Effectiveness with Sub-question Guidance}
\label{sec:train_policy_model}

In this part, we present the training details of the reasoner $\pi_{\theta_\text{R}}$.
First, to provide richer guidance for RLVR training, we use the trained decomposer to annotate the training data, supplying sub-question hints for each original question, before the beginning of the reasoner RLVR process. 
We then introduce an In-Context Distillation NLL Loss to guide the model’s exploration using the sub-question hints. 
After that, we provide the methods to integrate this loss function into the RLVR algorithms.

\paratitle{Annotating Training Data with Sub-question Hint.}
In Sec.~\ref{sec:train_decompoer}, we introduce the approach to train a decomposer only with the dataset containing questions and corresponding answers $\mathcal{D}=\{\langle x_i,y_i \rangle\}_{i=1}^n$.
This dataset will also be utilized in the RLVR process of the reasoner.
To provide additional guidance for the training procedure of the reasoner, we use the decomposer to decomposer each question in the training dataset and annotate it with a list of its potential sub-questions.
Formally, given the question $x_i$, the decomposer will decompose this question into several sub-questions $\mathcal{S}_i=\{s_{i,1},\dots,s_{i,p}\}$, and construct them into triples $\langle x_i, y_i, \mathcal{S}_i \rangle$.
After annotating all questions in the dataset $\mathcal{D}$, we construct the training dataset with sub-question hint using the above triples, \ie $\mathcal{D}_\text{SQ}=\{\langle x_i, y_i, \mathcal{S}_i \rangle\}_{i=1}^n$.
These sub-questions can be viewed as an idea of solving the corresponding question. 
After the reasoner sequentially addresses these sub-questions, it is likely to arrive at the solution to the original problem.

\paratitle{In-context Distillation Loss.}
In this part, we introduce \textit{in-context distillation loss} (IDL) to enhance the model’s capabilities after RLVR training with the guidance of the sub-question hint.
IDL can be regarded as a plug-and-play component that is compatible with various RLVR algorithms.
Specifically, given the question $x_i$ and its decomposed sub-questions $\mathcal{S}_i$, the reasoner generates multiple solutions $\hat{\mathcal{Y}'}=\{\hat{y}_{i,1}',\dots,\hat{y}_{i,n_\text{rollout}}'\}$.
With the guidance of sub-questions $\mathcal{S}_i$, the generated solutions $\hat{\mathcal{Y}}$ are more likely to contain the correct answer.
After generation, we verify the correctness of each response in $\hat{\mathcal{Y}}'$, obtaining the corresponding rewards $\{R_{i,1}',\dots,R_{i,n_\text{rollout}}'\}$.
Based on the rewards of the solutions, we can select the positive solutions to fine-tune the parameters of the reasoner $\pi_{\theta_\text{R}}$.
Since no sub-question hints are available during real testing, we train the model to generate solutions directly from the question, aiming to internalize the correct problem-solving strategies into the inner knowledge of the reasoner.
The loss function can be formulated as follows,
\begin{equation}
\small
    \mathcal{L}_\text{IDL}(\theta_\text{R})=-\frac{1}{N_\text{pos}}\sum_{j=1}^{n_\text{rollout}}\pi_{\theta_\text{R}}(\hat{y}_{i,j}'|x_i)\cdot \mathbb{I}[R_{i,j}'=1],~~N_\text{pos}=\sum_{j=1}^{n_\text{rollout}}\mathbb{I}[R_{i,j}'=1],
\end{equation}
where $\mathbb{I}[R_{i,j}'=1]$ is an indicator function, which returns $1$ when and only when $R_{i,j}'=1$ and returns $0$ for other situations.
By optimizing the reasoner’s parameters $\theta_\text{R}$ to minimize the loss function $\mathcal{L}_\text{IDL}$, the reasoner $\pi_{\theta_\text{R}}$ can learn more from these hints, enabling it to surpass its prior performance boundary.

\paratitle{Integrating IDL into RLVR Procedure.}
Following the process of vanilla RLVR training, for a given problem, we first have the reasoner generate multiple solutions $\hat{\mathcal{Y}}=\{\hat{y}_{i,1},\dots,\hat{y}_{i,n_\text{rollout}}\}$, and then compute the reward for each solution as well as the average reward $\bar{R}$.
For integrating the IDL function into the RLVR training process, the purpose is to provide auxiliary guidance for questions on which the reasoner is unable to obtain correct answers through the rollout process.
Following this motivation, we preset a threshold $k_1$ ($0<k_1<1$) such that when the average reward falls below this threshold (\ie $k_1>\bar{R}$), the IDL is applied. Otherwise, the IDL is not activated.
Moreover, since the reasoner $\pi_{\theta_\text{R}}$ is likely to follow the guidance to solve the question, the generated responses might show high similarity to each other.
Thus, to prevent overfitting, we collect the positive response generated under the sub-question guidance $\hat{\mathcal{Y}}'$, and retain at most $k_2 \times n_\text{rollout}$ ($0<k_2<1$) positive responses.
Formally, the set of the \textit{selected positive responses} can be denoted as 
\begin{equation}
\label{eq:process_y_with_sq}
\small
    \hat{\mathcal{Y}}_\text{Sel}'=\{\hat{y}_{i,1,\text{Sel}},\dots,\hat{y}_{i,q,\text{Sel}}\},~~q=\min(N_\text{pos}, k_2\times n_\text{rollout}).
\end{equation}
Furthermore, we observe that several generated solutions in $\hat{\mathcal{Y}}_\text{Sel}'$ are not rigorous, as they might fail to consider the scenarios not covered in the hints, thereby performing imprecise reasoning.
In this situation, directly fine-tuning the parameters of the reasoner on these imprecise data will hurt its performance.
Therefore, we appended different prompts to the original problem and used these prompts to segment the model's capabilities, as mentioned in previous work~\cite{thinkdial,prompt_engineering,prompt_format}.
The following formula presents the adjusted IDL,
\begin{equation}
\label{eq:idl}
\small
    \mathcal{L}_\text{IDL}'(\theta_\text{R})=-\frac{1}{|\hat{\mathcal{Y}}_\text{Sel}'|}\sum_{j=1}^{|\hat{\mathcal{Y}}_{\text{Sel}}'|}\pi_{\theta_\text{R}}(\hat{y}_{i,j,\text{Sel}}'|x_i,p).
\end{equation}
Based on the above discussion, we can rewrite the function of IDL and integrate it into the training process of RLVR, which can be formulated as follows,
\begin{equation}
\label{eq:a2d}
\small
    \mathcal{L}_\text{A$^2$D}(\theta_\text{R}) = - \mathcal{J}(\theta_\text{R}) + \alpha\mathcal{L}_\text{IDL}'(\theta_\text{R}) \cdot \mathbb{I}[k_1>\bar{R}],
\end{equation}
where $\mathcal{J}(\theta_\text{R})$ is the objective function of RLVR (Eq.~\ref{eq:grpo_loss_function}), and $\alpha$ is a hyperparameter to balance the effect of the above two objective functions.
\section{Experiments}
\label{sec:exp}

In this section, we conduct the experiment to evaluate the effectiveness of our approach and further analyze its features.
First, we introduce the details of the experimental settings (Sec.~\ref{sec:experiment_settings}).
Next, we present the results of the evaluation process (Sec.~\ref{sec:main_results}).
Finally, we present the detailed analysis of the feature of our approach (Sec.~\ref{sec:detailed_analysis}).

\subsection{Experimental Settings}
\label{sec:experiment_settings}

In this part, we introduce the datasets that are utilized in the training and evaluation process, the baseline approaches, and the implementation details of our experiments.

\paratitle{Datasets.}
We utilize the dataset proposed in previous work~\cite{still-3} as the training dataset for the decomposer and reasoner.
To perform the evaluation, we adopt eight mathematical benchmarks as the downstream tasks, including AIME24~\cite{aime24}, AIME25~\cite{aime25}, AMC23~\cite{amc23}, BeyondAIME~\cite{beyondaime}, MATH500~\cite{math500}, Minerva~\cite{minerva}, OlymMATH-Easy~\cite{olymmath}, and OlymMATH-Hard~\cite{olymmath}.

\paratitle{Baselines.}
To assess the effectiveness of our approach, we conduct several popular approaches as the baselines in our evaluation, including SFT-based and RLVR-based methods.
For SFT-based methods, we utilize seed-1.6 (no\_thinking mode)~\cite{seed1.6} to generate the solution and leverage the generated instances to perform SFT on the backbone models (\ie SFT w/ CoT).
Besides, given the sub-questions generated by the decomposer mentioned in our approach (Sec.~\ref{sec:train_decompoer}), we combine these generated sub-questions and the solutions mentioned above to conduct the SFT process (\ie SFT w/ CoT \& SQ).
For RLVR-based methods, we compare the performance of our approach with the previous work, \ie LUFFY~\cite{luffy} and Scaf-GRPO~\cite{scaf-grpo}.
Since these two approaches require an external teacher model to provide guidance, to conduct a fair comparison, we utilize our well-trained decomposer mentioned in Sec.~\ref{sec:train_decompoer} to provide the guidance.

\paratitle{Implementation Details.}
We adopt the Qwen2.5-7B-Instruct as the backbone model of the decomposer and employ GRPO~\cite{grpo} as the backbone RLVR algorithm for the whole training process.
To enhance the effectiveness, we follow the previous work~\cite{dapo} to set the clip ratio $\varepsilon_\text{low}=0.2$ and $\varepsilon_\text{high}=0.28$.
Moreover, we set the \texttt{max\_prompt\_length} and \texttt{max\_response\_length} as 2048 and 6144, and set the \texttt{train\_batch\_size} and \texttt{mini\_ppo\_batch\_size} as 128 and 32, respectively.
For each question, we perform $32$ times rollout process to conduct the experience for RLVR training, and utilize $1.0$ and $1.0$ as the \texttt{temperature} and \texttt{top\_p}.
For the evaluation procedure, we adopt the \texttt{temperature} as $1.0$ and \texttt{top\_p} as $1.0$ for the decoding process.
Besides, to reduce the influence of random values, we repeat the experiment 8 times and report the average accuracy.

\subsection{Main Results}
\label{sec:main_results}

\begin{table}[t]
    \centering
    \small
    \setlength{\tabcolsep}{3pt}
    \caption{Accuracy of different training approaches on mathematical tasks. BeAIME, OMATH-E, and OMATH-H denote BeyondAIME, OlymMATH-Easy, and OlymMATH-Hard, respectively. ``Avg.'' refers to the average accuracy of all tasks. The best is in bold and the second best is underlined.}
      \begin{tabular}{lccccccccc}
      \toprule
       & \textbf{AIME24} & \textbf{AIME25} & \textbf{AMC23} & \textbf{BeAIME} & \textbf{MATH500} & \textbf{Minerva} & \textbf{OMATH-E} & \textbf{OMATH-H} & \textbf{Avg.} \\
      \midrule
      \multicolumn{10}{c}{\textit{Qwen2.5-7B-Instruct}} \\
      Baseline & 9.6 & 6.3 & 47.5 & 3.4 & 70.8 & 22.9 & 4.8 & 2.1 & 20.9 \\
      + GRPO & \underline{15.0} & 8.3 & 52.8 & 5.3 & 73.3 & 24.8 & 4.6 & \underline{2.9} & 23.4 \\
      + GRPO w/ $n_\text{rollout}=64$ & 15.0 & 6.7 & 51.9 & 5.4 & 73.9 & \underline{25.3} & 4.6 & \textbf{3.0} & 23.2 \\
      + SFT w/ CoT & 7.1 & 2.9 & 41.3 & 2.8 & 63.6 & 24.4 & 2.9 & 1.4 & 18.3 \\
      + SFT w/ CoT \& SQ & 5.8 & 1.3 & 40.0 & 2.5 & 62.7 & 23.3 & 4.3 & 1.4 & 17.7 \\
      + LUFFY  & 10.4 & 8.8 & 47.2 & 3.9 & 69.2 & 22.4 & 4.4 & 2.8 & 21.1 \\
      + Scaf-GRPO & 14.6 & \underline{9.6} & \textbf{54.7} & \underline{5.5} & \underline{74.5} & 25.1 & \underline{5.6} & 2.8 & \underline{24.1} \\
      + A$^2$D (Ours) & \textbf{20.0} & \textbf{17.1} & \underline{53.4} & \textbf{6.6} & \textbf{75.6} & \textbf{28.9} & \textbf{8.4} & 1.8 & \textbf{26.5} \\
      \midrule
      \multicolumn{10}{c}{\textit{LLaMA3.2-3B-Instruct}} \\
      Baseline & 1.7 & 0.0 & 10.9 & 0.1 & 26.8 & 6.6 & 1.0 & 0.4 & 5.9 \\
      + GRPO & \underline{11.3} & 0.8 & 29.4 & 0.6 & \underline{52.5} & 13.4 & 2.5 & \underline{1.3} & 14.0 \\
      + GRPO w/ $n_\text{rollout}=64$ & 10.8 & 1.7 & 30.0 & 1.1 & \underline{52.5} & \underline{14.6} & \underline{2.8} & \underline{1.3} & \underline{14.4} \\
      + SFT w/ CoT & 2.1 & \underline{1.7} & 22.5 & 0.5 & 41.5 & 9.6 & 2.1 & 0.6 & 10.1 \\
      + SFT w/ CoT \& SQ & 3.8 & 0.4 & 21.9 & \underline{0.8} & 41.6 & 9.9 & 1.5 & 1.0 & 10.1 \\
      + LUFFY & 1.7 & 0.0 & 9.7 & 0.6 & 26.4 & 6.6 & 1.6 & 1.1 & 6.0 \\
      + Scaf-GRPO & \underline{11.3} & 0.8 & \underline{30.6} & \underline{0.8} & 51.2 & \textbf{15.3} & 2.6 & 0.9 & 14.2 \\
      + A$^2$D (Ours) & \textbf{13.8} & \textbf{2.1} & \textbf{30.9} & \textbf{2.0} & \textbf{52.6} & \textbf{15.3} & \textbf{3.1} & \textbf{1.6} & \textbf{15.2} \\
      \midrule
      \multicolumn{10}{c}{\textit{Qwen2.5-Math-7B-Instruct}} \\
      Baseline & 10.8 & 6.7 & 56.3 & 6.4 & 78.5 & 23.4 & 5.9 & 2.0 & 23.8 \\
      + GRPO & 10.4 & 8.8 & 58.1 & 8.3 & 79.7 & 24.6 & \textbf{8.7} & 2.0 & \underline{25.1} \\
      + A$^2$D (Ours) & \textbf{11.7} & \textbf{11.3} & \textbf{58.8} & \textbf{9.0} & \textbf{80.3} & \textbf{25.6} & \underline{8.4} & \textbf{3.0} & \textbf{26.0} \\
      \midrule
      \multicolumn{10}{c}{\textit{LLaMA3.1-8B-Instruct}} \\
      Baseline & 2.5 & \underline{0.0} & 13.8 & 0.1 & 33.3 & 10.9 & \underline{0.4} & 0.6 & 7.7 \\
      + GRPO & \underline{7.1} & \underline{0.0} & \underline{28.8} & \underline{0.8} & \underline{50.3} & \underline{19.6} & \textbf{1.6} & \underline{1.0} & \underline{13.7} \\
      + A$^2$D (Ours) & \textbf{8.3} & \textbf{1.3} & \textbf{30.6} & \textbf{1.4} & \textbf{50.6} & \textbf{19.9} & \textbf{1.6} & \textbf{1.6} & \textbf{14.4} \\
      \bottomrule
      \end{tabular}
      \label{tab:main_results}
\end{table}

For a more comprehensive evaluation, we use four backbone models, train them with our method and baseline approaches, and present the evaluation results in Table~\ref{tab:main_results}.

First, from the experimental results, we observe that our method A$^2$D consistently yields substantial improvements across different backbone models and achieves superior performance on mathematical reasoning tasks.
Compared with both the SFT-based (\ie SFT w/ CoT and SFT w/ CoT \& SQ) and RLVR-based (\ie GRPO, LUFFY, and Scaf-GRPO) baseline approaches, the models trained with our method also surpass these baselines and achieve better performance.
The experimental results demonstrate that our method is effective across models of different families, scales, and capability levels.

Second, our method A$^2$D demonstrates the robustness across various models. 
During the training of the Decomposer, we use Qwen2.5-7B-Instruct as the proxy reasoner.
Once the Decomposer is trained, its decomposed sub-questions can assist not only the RLVR training of Qwen2.5-7B-Instruct but also that of Qwen2.5-Math-7B-Instruct and other models from the LLaMA family, yielding performance improvements for these models.
This indicates that the trained Decomposer can guide the training of various models without requiring specific training for different reasoners, demonstrating that our method A$^2$D is both efficient and effective.

Third, comparing the two SFT-based training methods, we find that adding decomposer-generated sub-questions to the training data does not improve model performance.
Instead, it even leads to performance degradation in certain scenarios, \eg AIME25 and AMC23.
The underlying reasons are twofold. 
First, learning to decompose the original question may not directly aid in solving it, as these are two dissimilar abilities. 
Second, although the decomposer-generated sub-questions can stimulate exploration that occasionally leads to the correct answer, they do not necessarily represent a complete reasoning process and may overlook certain cases. 
For these reasons, directly learning the decomposer-generated sub-questions does not contribute to improving the model’s reasoning ability.
Our method effectively avoids this issue by leveraging the decomposed sub-questions through the IDL function, thereby providing richer guidance for the RLVR process.

Finally, directly training the model using teacher-generated solutions that do not contain the thinking process (\ie SFT w/ CoT and LUFFY) is unlikely to yield improvements and may even impair the model’s reasoning ability.
A possible explanation for this phenomenon is that these backbone models have already learned a substantial amount of instructions during post-training. Further training on such data may lead to overfitting, thereby degrading their performance on the test set.
Compared to SFT with CoT, LUFFY introduces data distilled from the teacher model only when the model struggles to answer a question, reinforcing knowledge that the model has not yet mastered and thereby partially alleviating the overfitting caused by learning from teacher data.
However, this approach requires the introduction of an external model and relies on the teacher model possessing strong capabilities to produce correct and complete reasoning processes.

\subsection{Detailed Analysis}
\label{sec:detailed_analysis}

To further understand the feature of our approach, in this part, we present the detailed analysis, including the ablation study (Sec.~\ref{sec:ablation_study}), the effectiveness of integrating A$^2$D into different RLVR algorithms (Sec.~\ref{sec:other_rlvr}), effects of different utilization for the generated sub-questions (Sec.~\ref{sec:concatenating_subquestion}), the statistical analysis of the sub-questions from the decomposer (Sec.~\ref{sec:statistical_analysis}), and the case study of the generated sub-questions 
(Appendix~\ref{sec:case_study}).

\subsubsection{Ablation Study}
\label{sec:ablation_study}

\begin{table}[t]
    \centering
    \small
    \caption{Results of the ablation study. We first present the ablation study about the training process of the decomposer, and then provide the analysis of the effectiveness of each module in our approach. The best is in bold.}
      \begin{tabular}{lccccc}
      \toprule
       & \textbf{AIME24} & \textbf{AIME25} & \textbf{MATH500} & \textbf{Minerva} & \textbf{Avg.} \\
      \midrule
      \multicolumn{6}{c}{\textit{Pass@k Performance of the Decomposer}} \\
      A$^2$D (Ours) & \textbf{20.8} & \textbf{18.3} & \textbf{82.9} & \textbf{30.7} & \textbf{38.2}  \\
      \quad w/o Format Reward w/ Pass@k Reward & 15.8 & 16.7 & 83.2 & 30.4 & 36.5  \\
      \quad w/o Format Reward w/ Pass@1 Reward & 16.3 & 14.9 & 81.5 & 28.5 & 35.3  \\
      \quad w/o RLVR & 17.1 & 13.0 & 78.6 & 28.1 & 34.2  \\
      \midrule
      \multicolumn{6}{c}{\textit{Pass@1 Performance of the Reasoner}} \\
      A$^2$D (Ours) & \textbf{20.0} & \textbf{17.1} & \textbf{75.6} & \textbf{28.9} & \textbf{35.4} \\
      \quad w/o Removing Guidance & 15.0 & 9.6 & 74.5 & 25.0 & 31.0  \\
      \quad w/o Selection & 16.3 & 11.7 & 73.7 & 26.6 & 32.1 \\
      \quad w/o Selection w/o Diversity Prompt & 13.3 & 10.0 & 69.6 & 22.6 & 28.9 \\
      \quad w/o Sub-question Guidance & 15.0 & 8.3 & 73.3 & 24.8 & 30.4 \\
      \bottomrule
      \end{tabular}
      \label{tab:ablation}
\end{table}

To assess the effectiveness of each module in the training process of A$^2$D, we conduct the ablation study and present the results in Table~\ref{tab:ablation}.

\paratitle{Ablation Study about Training Process of Decomposer.}
To evaluate the capability of the decomposer, we concatenate the sub-questions that it generates with the original question as a combined prompt, and then ask the LLM (\ie Qwen2.5-7B-Instruct) to produce an answer. We report the Pass@k score in the top part of Table~\ref{tab:ablation}.
Based on the evaluation results, the Format Reward and Pass@k Reward both contribute to the enhancement of the capacities of decomposer.
Format Reward teachs decomposer to generate a formal list of subproblems, and Pass@k Reward helps it to present the sub-problems that are more likely to guide the reasoner to arrive at the correct answer.
By training with the above two rewards, the decomposer can better assist the subsequent RLVR training process of the reasoner, thereby improving the training effect.

\paratitle{Ablation Study about RLVR with Sub-question Guidance.}
In the bottom part of the Table~\ref{tab:ablation}, we present the results of removing Selection and Deversirty Prompt mechanism of our approach.
We can observe the decrease of the model's performance without these two modules.
For problems that the model is already capable of solving through its own exploratory reasoning, it is unnecessary to introduce additional external guidance, so as to avoid potential conflicts between such guidance and the model’s inherent problem-solving trajectory.
For problems that require external guidance, training with diverse prompts can mitigate the influence of external knowledge on the model, thereby improving its overall performance.

\subsubsection{Adaptation to Other RLVR Algorithms}
\label{sec:other_rlvr}

\begin{table}[t]
    \centering
    \small
    \setlength{\tabcolsep}{3.5pt}
    \caption{Accuracy of Qwen2.5-7B-Instruct trained through different approaches on mathematical tasks. BeAIME, OMATH-E, and OMATH-H denote BeyondAIME, OlymMATH-Easy, and OlymMATH-Hard, respectively. ``Avg.'' refers to the average accuracy of all tasks. The best is in bold.}
      \begin{tabular}{lccccccccc}
      \toprule
       & \textbf{AIME24} & \textbf{AIME25} & \textbf{AMC23} & \textbf{BeAIME} & \textbf{MATH500} & \textbf{Minerva} & \textbf{OMATH-E} & \textbf{OMATH-H} & \textbf{Avg.} \\
      \midrule
      Baseline & 9.6 & 6.3 & 47.5 & 3.4 & 70.8 & 22.9 & 4.8 & 2.1 & 20.9 \\
      \midrule
      + GRPO & 15.0 & 8.3 & 52.8 & 5.3 & 73.3 & 24.8 & 4.6 & 2.9 & 23.4 \\
      + A$^2$D (Ours) & 20.0 & 17.1 & 53.4 & 6.6 & 75.6 & 28.9 & 8.4 & 1.8 & \textbf{26.5} \\
      \midrule
      + RLOO & 12.9 & 7.1 & 52.5 & 4.6 & 74.9 & 25.2 & 5.9 & 2.8 & 23.2 \\
      + A$^2$D (Ours) & 17.9 & 12.5 & 55.0 & 5.8 & 75.2 & 26.8 & 6.4 & 2.3 & \textbf{25.2} \\
      \midrule
      + REINFORCE++ & 14.2 & 8.3 & 50.6 & 3.6 & 73.6 & 26.1 & 3.6 & 2.1 & 22.8 \\
      + A$^2$D (Ours) & 14.6 & 10.4 & 52.8 & 5.6 & 72.7 & 26.7 & 4.9 & 2.5 & \textbf{23.8} \\
      \bottomrule
      \end{tabular}
      \label{tab:other_rlvr}
\end{table}

Our method introduces the IDL function to guide the model in leveraging hints during RLVR training to further enhance its capabilities. 
Theoretically, our approach is orthogonal to existing RLVR methods and can be combined with other RLVR algorithms to improve their training effectiveness.
To further validate our approach experimentally, we collected commonly used RLVR training algorithms (\eg GRPO~\cite{grpo}, RLOO~\cite{rloo}, and REINFORCE++~\cite{reinforce++}) and combined A$^2$D with these methods. The results are presented in Table~\ref{tab:other_rlvr}.

From the experimental results, we observe that traditional RLVR algorithms can enhance the model’s reasoning ability. Building on these algorithms, our A$^2$D can further improve the model’s capabilities and achieve better performance on downstream tasks.
For more challenging tasks (\eg AIME25 and BeyondAIME), vanilla RLVR struggles to yield improvements, as the model’s limited capabilities make it difficult to discover correct solutions through self-exploration and thus produce accurate answers.
In such cases, the decomposer’s breakdown of the original problem can provide the reasoner with novel problem-solving strategies, extending its capability boundaries and enabling it to tackle these more complex tasks during the RLVR training process.
Through our analysis, we find that A$^2$D can be adapted to various RLVR algorithms, effectively enhancing the model’s capabilities.

\subsubsection{Effect of Concatenating Decomposed Sub-questions into Prompt}
\label{sec:concatenating_subquestion}

\begin{figure}[t]
    \centering
    \subfloat[Pass@1 Performance without Training.]{
        \includegraphics[width=0.48\linewidth]{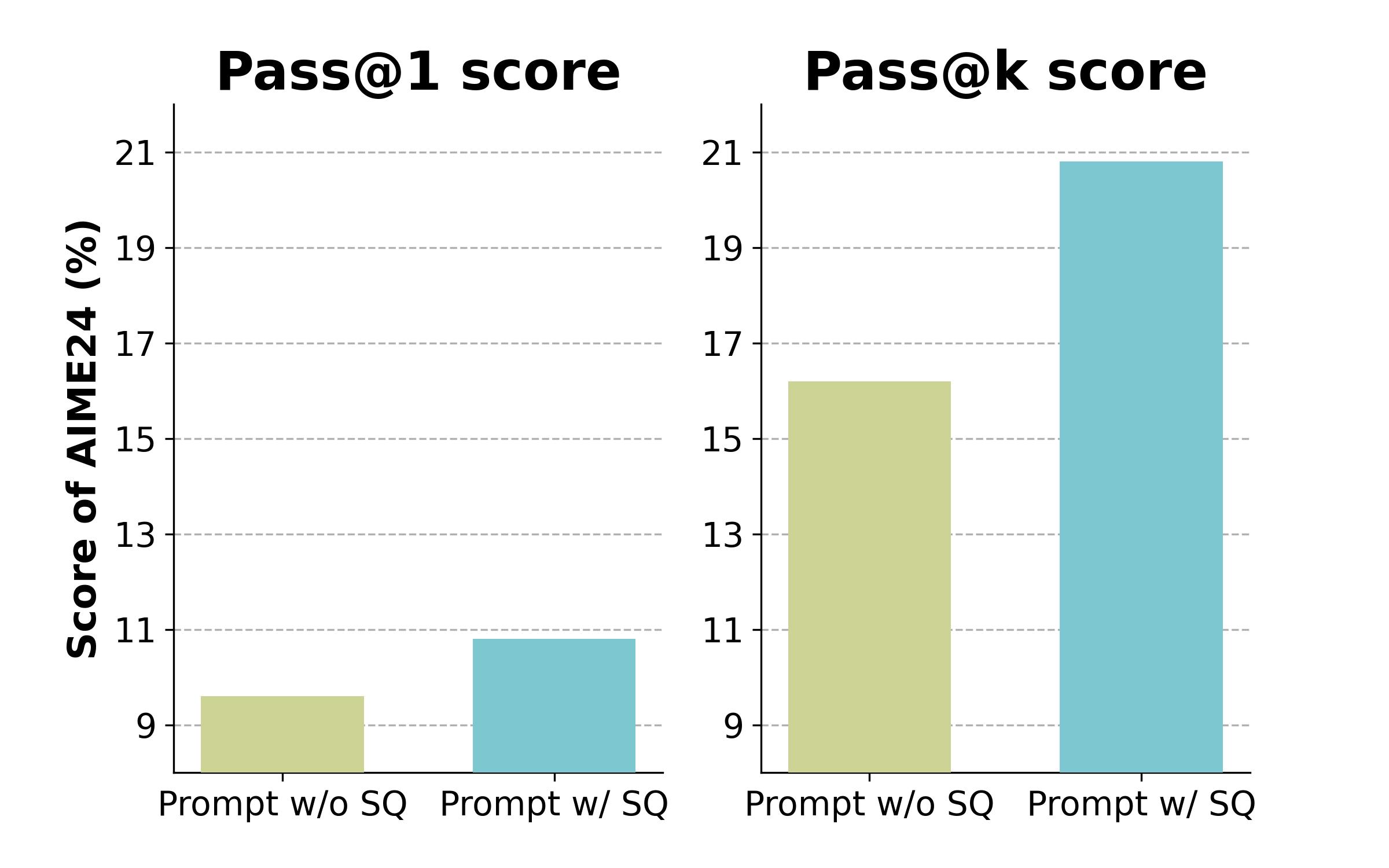}
        \label{fig:concatenating_subquestion_notrain}
    }
    \subfloat[AIME24 Performance of AIME24 during GRPO.]{
        \includegraphics[width=0.48\linewidth]{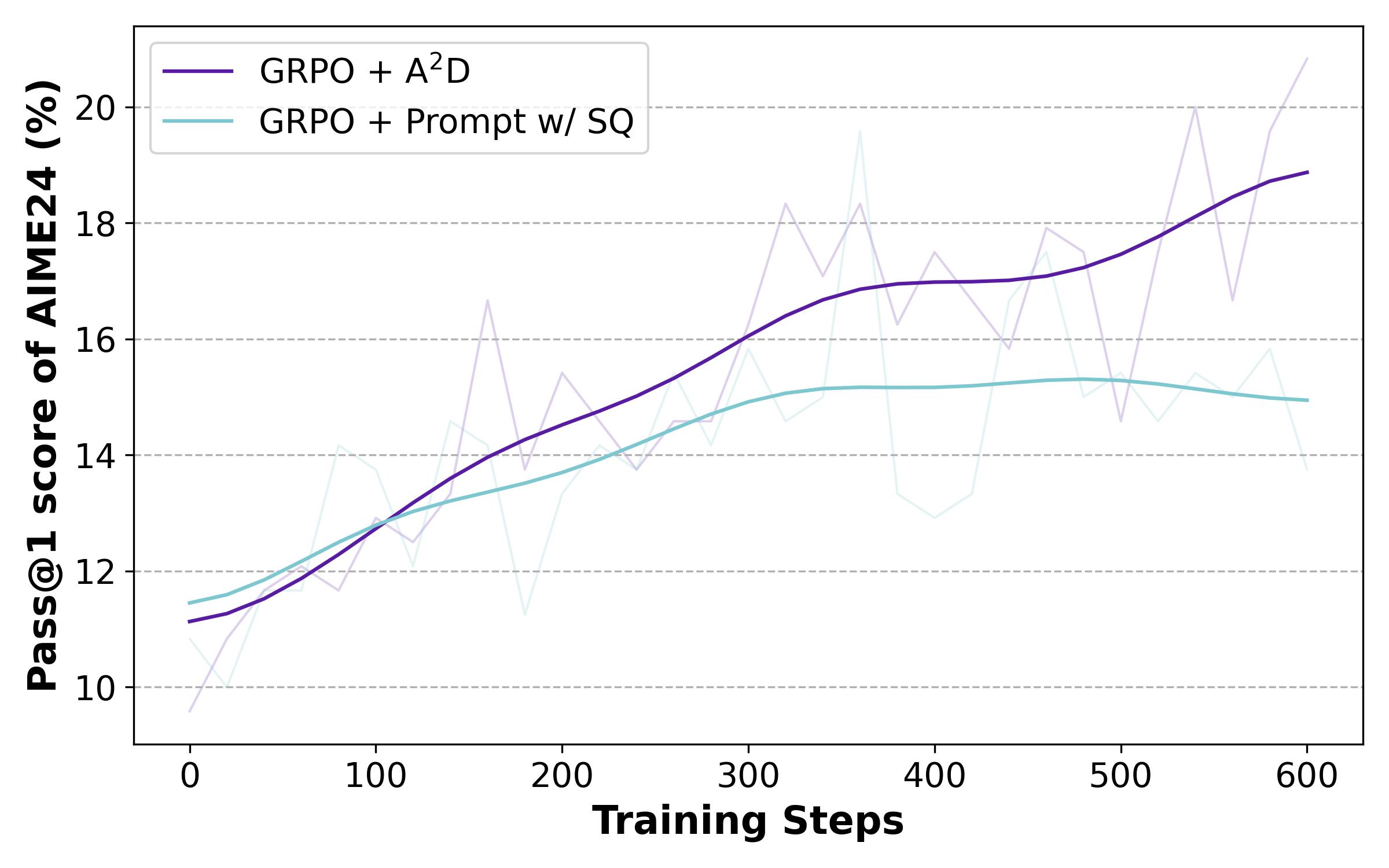}
        \label{fig:sconcatenating_subquestion_Adaptivetrain}
    }
    \caption{Performance comparison of Qwen2.5-7B-Instruct of whether sub-questions are included in the prompt.}
    \label{fig:concatenating_subquestion}
\end{figure}

As mentioned in Sec.~\ref{sec:train_decompoer}, we can train a decomposer through RLVR, enabling it to break down the original question into simpler sub-questions that help the reasoner perform more effective reasoning.
In fact, the trained decomposer can not only be used to decompose questions in the training set, but also to add sub-question hints to test-time queries, thereby assisting the model in solving downstream tasks.
To evaluate the effectiveness of different methods, we conduct experiments and present the results in Fig.~\ref{fig:sconcatenating_subquestion_Adaptivetrain}.

In Fig.~\ref{fig:concatenating_subquestion_notrain}, we observe that adding sub-questions generated by the decomposer into the prompt (\ie Prompt w/ SQ) leads to improvements in both the Pass@1 score and the Pass@k score.
This phenomenon indicates that decomposing the original problem can guide large models to explore and reason more effectively, resulting in improved performance. 
This also serves as one of the key motivations behind our method.

Moreover, we added sub-question hints to both the training and test sets and trained the reasoner using GRPO on this augmented data \ie GRPO + Prompt w/ SQ.
In Fig.~\ref{fig:sconcatenating_subquestion_Adaptivetrain}, we compare its performance with our approach (\ie GRPO + A$^2$D).
We observe that in the early stages of training (first $100$ steps), GRPO + Prompt w/ SQ achieves better performance. This is because the sub-question hints can directly provide the reasoner with solution strategies, leading to improved reasoning performance.
However, as training progresses, the model becomes overly reliant on the sub-question hints provided by the decomposer for reasoning, which diminishes its exploratory capability and prevents further improvement in reasoning performance.
In contrast, during the training of A$^2$D, the model uses hints only when the question is sufficiently difficult, while relying on self-exploration in all other cases. 
As a result, A$^2$D does not lead to a rapid decline in the model’s exploratory ability and enables the model to achieve better performance on downstream tasks through RLVR.

\subsubsection{Statistical Analysis of the Generated Sub-questions from A$^2$D}
\label{sec:statistical_analysis}

\begin{table}[t]
    \centering
    \small
    \caption{The statistical information of the generated sub-questions from A$^2$D.}
      \begin{tabular}{lcccc}
      \toprule
       & \textbf{AIME24} & \textbf{AIME25} & \textbf{MATH500} & \textbf{Minerva} \\
      \midrule
      \multicolumn{5}{c}{\textit{Number of Sub-questions}} \\
      Average value & 2.07 & 2.1 & 2.02 & 1.99  \\
      Standard deviation & 0.25 & 0.4 & 0.15 & 0.28  \\
      Maximum value & 3.00 & 4.00 & 3.00 & 5.00  \\
      Minimum value & 2.00 & 2.00 & 1.00 & 1.00  \\
      \midrule
      \multicolumn{5}{c}{\textit{Number of Tokens of Generated Sub-questions}} \\
      Average value &  60.37 & 58.33 & 50.08 & 53.01 \\
      Standard deviation & 23.93 & 22.19 & 15.99 & 17.33  \\
      Maximum value &  115.00 & 154.00 & 130.00 & 167.00 \\
      Minimum value & 31.00 & 36.00 & 26.00 & 16.00  \\
      \midrule
      \multicolumn{5}{c}{\textit{Contained Content}} \\
      Whether contain ``\verb|\boxed{}|'' & \ding{55} & \ding{55} & \ding{55} & \ding{55} \\
      Whether contain ``\verb|Answer:|'' & \ding{55} & \ding{55} & \ding{55} & \ding{55} \\
      Whether contain ``\verb|Answer is|'' & \ding{55} & \ding{55} & \ding{55} & \ding{55} \\
      \bottomrule
      \end{tabular}
      \label{tab:statistical_information}
\end{table}

To further understand the feature of the decomposer, we collect the generated sub-questions from decomposer for four dataset, \ie AIME24, AIME25, MATH500, and Minerva, and present the statistical information about the generated sub-questions in Table~\ref{tab:statistical_information}.

First, by examining the number of sub-questions produced by the decomposer and the number of generated tokens, we observe that both remain at relatively low levels.
Specifically, on average, a question can be decomposed into two sub-questions, and describing these sub-questions requires only about 60 tokens.
This phenomenon indicates that the sub-questions produced by the decomposer operate at a relatively high level of abstraction and can be regarded as a coarse-grained reasoning process rather than a step-by-step solution.

Moreover, we find that the responses generated by the decomposer do not contain keywords related to the final answer, indicating that the decomposer provides problem-solving guidance without revealing the solution itself, thereby not constraining the reasoner’s reasoning and exploratory process.
Such a decomposer can provide the reasoner with guidance and inspiration while retaining a certain degree of exploratory capability, which is precisely what we aim for.
The decomposer does not engage in overly detailed reasoning that would excessively reinforce exploitative behavior and consequently diminish its exploratory capacity.

\subsubsection{Case Study of the Generated Sub-questions from Different Models}
\label{sec:case_study}

\begin{figure}[h]
    \centering
    \includegraphics[width=1\linewidth]{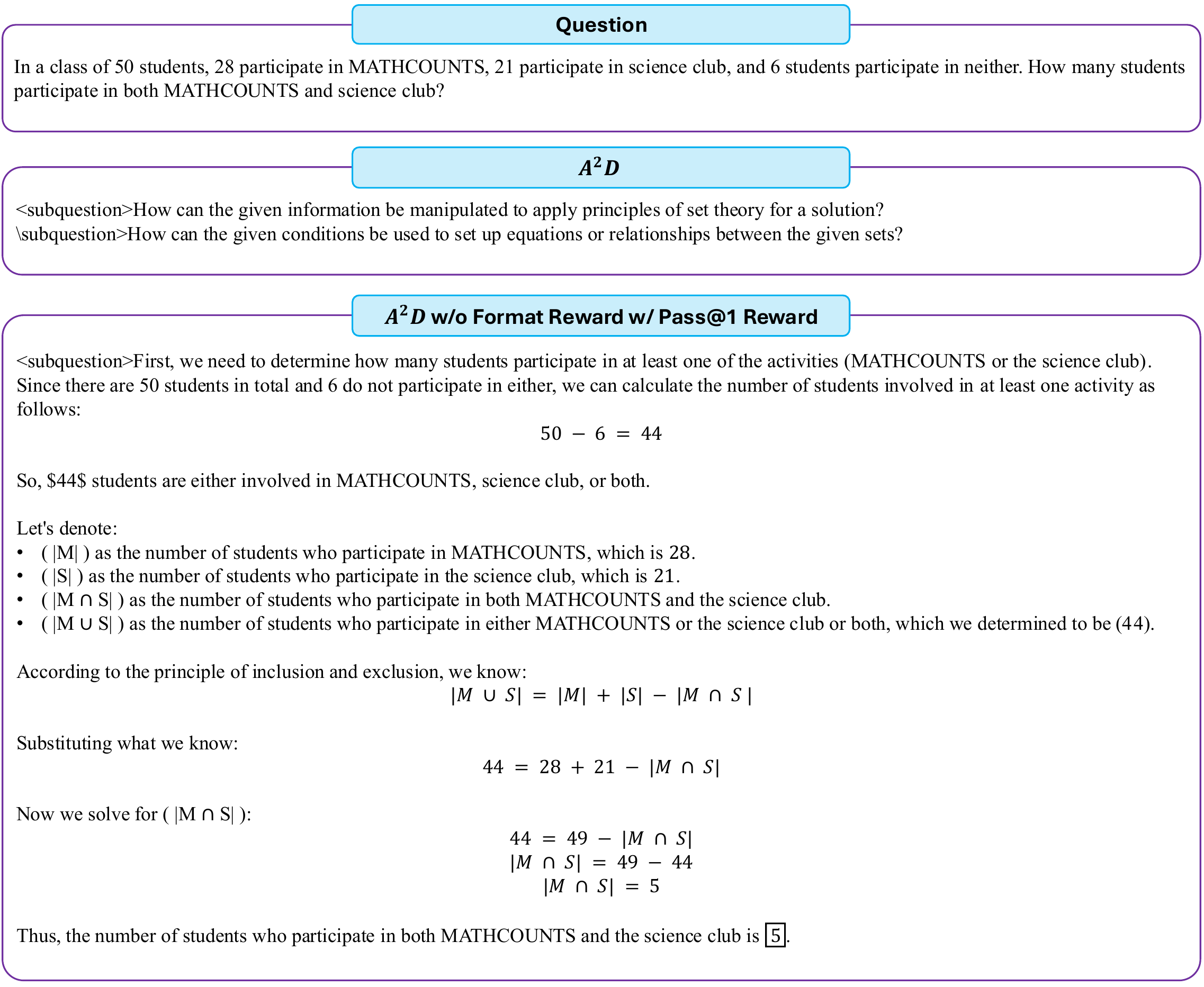}
    \caption{An example of the generated sub-question from models trained through different methods.}
    \label{fig:case_study}
\end{figure}

In Fig.~\ref{fig:case_study}, we present the case study of the generated sub-questions of different models, \ie our method (\ie A$^2$D) and the approach that removes the Format Reward while using the Pass@1 Reward to replace the Pass@k Reward (\ie A$^2$D w/o Format Reward w/ Pss@1 Reward).

The example problem involves sets and requires knowledge of intersections and unions.
The model trained with our method decomposes this problem into two sub-questions: first, it identifies that the problem involves sets, guiding the reasoner to select relevant knowledge for solving the problem; then, it guides the reasoner to apply this knowledge to the specific question.
The Decomposer performs a relatively coarse-grained decomposition of the problem, which can inspire the reasoner to explore while preventing over-reliance on the hints. This phenomenon reflects the effect of using the Pass@k Reward during training.

In contrast, the model trained by removing the Format Reward and using the Pass@1 Reward instead of the Pass@k Reward produces more concrete content, with the sub-questions degenerating into step-by-step reasoning solutions.
This phenomenon indicates that high-level, coarse-grained hints can guide the model’s exploration, thereby improving its Pass@k performance.
In contrast, fine-grained hints better enhance the model’s exploitative capability, leading to higher Pass@1 performance.
The analysis of the model’s generated content validates the rationale of our training method.
\section{Related Work}

\subsection{Improving Exploration Ability of LLMs via Value-based Guidance}
RLVR is a popular approach to achieve test-time scaling, which utilizes reward scores to supervise LRMs~\cite{openai-o1,deepseek-r1,kimi-k1.5}.
During the RLVR process, the reward design~\cite{Arnal-arxiv-2025-Asymmetric,Su-arxiv-2025-Crossing}, advantage shaping~\cite{still-4,passk_training,entropyadv}, and sampling mechanism~\cite{forking_token,Zhu-arxiv-2025-Surprising} are leveraged to unlock the potential of LRMs, balancing their exploration and exploitation that guides them to find the global optimal.
In addition to enabling the model to perform exploration via one-step generation, previous works~\cite{tot,got,alphamath,still-1} have employed reward models to provide supervision signals for reasoning steps, thereby facilitating more robust reasoning.
Besides, existing work~\cite{llamaberry} also adopts outcome-level reward models to guide the reasoning processes of the models.
Given that models can effectively solve simple problems but struggle with multi-hop reasoning tasks~\cite{scaf-grpo,Xue-nips-2024-Decompose,Peng-acl-2024-Chain}, this work focuses on how to guide the model to transfer its ability to solve subproblems toward solving more complex problems, thereby enhancing the effectiveness of RLVR. In principle, our approach is orthogonal to previous work and can be effectively combined with those methods.

\subsection{Test-time Scaling via Natural Language Guidance}
Beyond the score-based supervision, natural language guidance is also leveraged to enhance the reasoning performance of LRMs.
Supervised finetuning (SFT) is widely used on LRMs to elicit their inner knowledge and improve their exploration ability~\cite{deepseek-r1,llama-nemotron}.
The training instances that contains the complex actions for the SFT process are usually distilled from the powerful teacher models~\cite{openr1,open-thought}, \eg GPT-5 or Gemini-2.5-pro.
Moreover, previous studies have shown that responses from the teacher model~\cite{luffy,scaf-grpo} and self-reflection~\cite{Sareen-arxiv-2025-Putting,pag} can be used to elicit the reasoning ability of LRMs, in the RLVR process.
Besides, parallel reasoning enhances the performance of LRMs by enabling the model to explore multiple reasoning paths and obtain the final answer based on the previous generation~\cite{Qi-arxiv-2025-Learning,Snell-arxiv-2024-scaling,learning_from_peers}, which is an alternative way of leveraging natural language guidance.
In this work, we investigate the use of the tips (\ie the sub-problems derived from the decomposition of the original problem) during training to aid the model in accurately solving these problems and to activate its internal knowledge.

\subsection{Compositional Generalization Ability of LLMs}
Compositional generalization ability refers to the ability to combine known simple knowledge to solve more complex problems~\cite{Liu-nips-2020-Compositional,An-acl-2023-How}.
However, the compositional generalization ability of LLMs is limited~\cite{Andreas-cvpr-2016-Neural,Sakai-acl-2025-Revisiting,Yang-naacl-2024-Exploring}.
Specifically, the model has acquired some basic knowledge, but it still cannot combine these related pieces of knowledge to solve more complex tasks, \eg puzzle tasks~\cite{passk_training,enigmata} and question answering tasks~\cite{search-r1,hotpotqa}.
To improve such an ability of LLMs, previous studies usually adopt several simple problems to synthesize the hard multi-hop problems and utilize these problems to perform training process~\cite{fx_gx_fgx,omega}, or utilize multi-turns reasoning solve the complex problems step-by-step~\cite{Jiang-www-2025-Retrieve,Zhu-acl-2025-Mitigating}.
The compositional generalization ability is vital for the self-evolution of LLMs~\cite{Sakai-acl-2025-Revisiting,Yang-naacl-2024-Exploring}, such as RLVR, where LLMs should explore the potential solutions to the given question based on their knowledge and learn from these experiences~\cite{scaf-grpo,stephint}.
Inspired by this, in this work, we focus on teaching LLMs to combine the knowledge it has learned to answer complex questions in RLVR, which can increase the efficiency of the model’s exploration, thereby improving the training effectiveness.

\section{Conclusion}

In this work, we proposed A$^2$D, adaptive ability decomposing for enhancing the effectiveness of the RLVR process.
Concretely, we first trained a decomposer to decompose the complex questions into several sub-questions, and then utilized these generated sub-questions to assist the RLVR procedure of the reasoner.
With the additional information, the reasoner can explore more effectively and further enhance its reasoning ability during the RLVR process.
To better understand the features of the decomposer, we conducted a detailed analysis of its training approach and the behaviours after training.
Through our experiments, we find that abstract and coarse-grained hints, \eg sub-question guidance, are beneficial for enhancing the model’s exploration ability, whereas concrete and fine-grained prompts are more effective for improving the model’s exploitation ability.
Our work investigated how to guide the model toward more effective exploration.

In future work, the compatibility between the decomposer and the reasoner is a promising direction for future research. 
As the reasoner’s capability improves, the style and granularity of the hints provided by the decomposer should also evolve accordingly.
Furthermore, whether our algorithm can be extended to online settings is also an important question for future exploration. While AI assistants help humans solve problems, they must also continuously evolve and improve based on real feedback provided by users.


\clearpage

\bibliographystyle{plainnat}
\bibliography{main}

\clearpage

\beginappendix

\section{Prompt Templates}

In this part, we introduce the details of the prompt templates that are utilized in out experiments.

Below is the prompt used to train the decomposer model, where the placeholder ``\{QUESTION\}'' is replaced with the corresponding question $x_i$.
\begin{tcolorbox}[
    colframe=takeaway,
    colback=white,
    coltitle=takeawayTitle,
]
\textcolor{takeawayTitle}{\textbf{Prompt for Decomposer}}

\{QUESTION\}

Please reason step-by-step and decompose this question into several sub-questions. You only provide the sub-questions and do not provide the any details reasoning steps and solution. Use the format ``<subquestion>'' and ``<\verb|\|subquestion>'' to identify each sub-question.
\end{tcolorbox}

Below is the prompt used to guide the reasoner to perform reasoning, which is employed in the objective function $\mathcal{J}(\theta_R)$ in Eq.~\ref{eq:grpo_loss_function}. The placeholder ``\{\texttt{QUESTION}\}'' is replaced with the corresponding question $x_i$.
\begin{tcolorbox}[
    colframe=takeaway,
    colback=white,
    coltitle=takeawayTitle,
]
\textcolor{takeawayTitle}{\textbf{Vanilla Prompt for Reasoner}}

\{QUESTION\} \\
Please reason step by step, and put your final answer within \verb|\\boxed{}|.
\end{tcolorbox}

Below is the prompt used by the reasoner for inference under sub-question guidance. The placeholder ``\{\texttt{QUESTION}\}'' is replaced with the corresponding original question $x_i$, and ``\{\texttt{LIST OF SUB-QUESTIONS}\}'' is replaced with the decomposed sub-questions $\mathcal{S}_i$.
\begin{tcolorbox}[
    colframe=takeaway,
    colback=white,
    coltitle=takeawayTitle,
]
\textcolor{takeawayTitle}{\textbf{Prompt with Sub-question Guidance for Reasoner}}

\{\texttt{QUESTION}\} \\\\
Now, I give you some tips and you can refer to the provided tips to solve the problem.
Tips:\\
\{\texttt{LIST OF SUB-QUESTION}\}\\\\
Please reason step by step, and put your final answer within \verb|\\boxed{}|.
\end{tcolorbox}

Below is the prompt used during IDN Loss training, and the placeholder ``{QUESTION} is replaced with the corresponding question.
\begin{tcolorbox}[
    colframe=takeaway,
    colback=white,
    coltitle=takeawayTitle,
]
\textcolor{takeawayTitle}{\textbf{IDN Loss Prompt for Reasoner}}

\{\texttt{QUESTION}\}\\\\
You can try to reason step by step through different approaches, and try to explore the correct the solution. You should put your final answer within \verb|\\boxed{}|.
\end{tcolorbox}

\section{Details of the Format Reward for Decomposer Training}
\label{app:details_format_reward}

We enforce a series of constraints to verify whether the model responses follow the required format.
\begin{itemize}
    \item The model’s generated response is required to begin with the ``<subquestion>'' tag.
    \item Each sub-question is not allowed to contain more than one ``<subquestion>'' tag.
    \item The model’s generated response must contain more than $10$ characters.
\end{itemize}

\section{Pseudo Code of A$^2$D}
In Algorithm~\ref{code:a2d}, we present the pseudocode the our A$^2$D to better demonstrate its training pipeline.

\begin{algorithm*}[h]
\small
\caption{The Pseudo Code for A$^2$D.}
\label{code:a2d}
\SetKwInOut{Input}{Input}
\SetKwInOut{Output}{Output}

\Input{A backbone model $\pi_\theta$ and the training dataset $\mathcal{D}=\{(x_i,y_i)\}_{i=1}^{n}$.}
\Output{A powerful reasoner $\pi_{\theta_\text{R}}$.}

\BlankLine
\BlankLine
\texttt{\# (1) Training Decomposer} $\pi_{\theta_\text{D}}$ \texttt{via RLVR.} \\
Initialize the decomposer $\pi_{\theta_\text{D}}$ and proxy reasoner using the backbone model $\pi_{\theta}$. \\
\For{each training step}{
    Prompt the decomposer $\pi_{\theta_\text{D}}$ to decompose the questions in the training dataset. \\
    Compute the quality reward of the decomposed question $R_\text{Q}$ by the proxy reasoner. \\
    Compute the format reward of the decomposed question $R_\text{F}$ by rules in Appendix~\ref{app:details_format_reward}. \\
    Utilize Eq.~\ref{eq:grpo_loss_function} to optimize the decomposer $\pi_{\theta_\text{D}}$ based on the reward $R = R_\text{Q}\times R_\text{F}$. \\
}
Obtain a decomposer $\pi_{\theta_\text{D}}$, which can decompose the complex question into several simpler sub-questions. \\

\BlankLine
\BlankLine
\texttt{\# (2) Decompose the Complex Questions.}\\
\For{each instance $\langle x_i, y_i \rangle$ in training dataset $\mathcal{D}$} {
    Use decomposer $\pi_{\theta_\text{D}}$ to decompose the question $x_i$, obtaining a list of sub-questions $\mathcal{S}_i$. \\
    Add the triple $\langle x_i, y_i, \mathcal{S}_i \rangle$ into the training dataset of the reasoner $\mathcal{D}_\text{SQ}$. \\
}
Obtain the training dataset of the reasoner $\mathcal{D}_\text{SQ}=\{\langle x_i, y_i, \mathcal{S}_i \rangle\}_{i=1}^{n}$. \\

\BlankLine
\BlankLine
\texttt{\# (3) Training Reasoner} $\pi_{\theta_\text{R}}$ \texttt{through RLVR with Sub-question Guidance.} \\
Initialize the reasoner $\pi_{\theta_\text{R}}$ using the backbone model $\pi_{\theta}$. \\
\For{each instance $\langle x_i, y_i, \mathcal{S} \rangle$ in training dataset $\mathcal{S}_\text{SQ}$}{
    Prompt the reasoner $\pi_{\theta_\text{R}}$ to generate the solutions $\hat{\mathcal{Y}}$, based on the question $x_i$. \\
    Compute the loss function of RLVR $\mathcal{J}(\theta_\text{R})$ using Eq.~\ref{eq:grpo_loss_function}, based on $\hat{\mathcal{Y}}$.
    \BlankLine
    
    Prompt the reasoner $\pi_{\theta_\text{R}}$ to generate the solutions $\hat{\mathcal{Y}}'$, based on the question $x_i$ and the sub-questions $\mathcal{S}_i$. \\
    Process the generated solutions $\hat{\mathcal{Y}}'$ through Eq.~\ref{eq:process_y_with_sq}. \\
    Compute the in-context distillation loss $\mathcal{L}'_\text{IDL}(\theta_\text{R})$ using Eq.~\ref{eq:idl}, based on $\hat{\mathcal{Y}}'$.
    \BlankLine

    Utilize Eq.~\ref{eq:a2d} to optimize the reasoner $\pi_{\theta_\text{R}}$ based on $\mathcal{J}(\theta_\text{R})$ and $\mathcal{L}'_\text{IDL}(\theta_\text{R})$. \\
}
Obtain the powerful reasoner $\pi_{\theta_\text{R}}$.

\end{algorithm*}

\end{document}